# ToxLens: A Reproducible Graph-Learning Framework for Leakage-Aware, Uncertainty-Calibrated Molecular Toxicity Prediction

Magnus H. Strømme[1], Alex G. C. de Sá[1,2,*], David B. Ascher[1,2,*]

[1]The Australian Centre for Ecogenomics, School of Chemistry and Molecular Biosciences, The University of Queensland, Brisbane, Queensland, 4072, Australia

[2]Computational Biology and Clinical Informatics, Baker Heart and Diabetes Institute, Melbourne, Victoria, 3004, Australia

[*]D.B.A., tel. +61 7 3365 3891, email d.ascher@uq.edu.au; or A.G.C.d.S., email a.desa@uq.edu.au.

**Abstract.** Molecular toxicity prediction is increasingly used to prioritise compounds before experimental testing, but conventional benchmark performance can overstate practical utility when structurally related molecules occur across training and test folds. We introduce ToxLens, a reproducible multi-task graph-learning framework for 11 toxicity endpoints spanning Ames mutagenicity, acute oral toxicity, hERG inhibition, and Tox21 nuclear-receptor and stress-response assays. The workflow combines conservative chemical curation, sphere-exclusion filtering, a leakage-aware UMAP-HDBSCAN split, parallel graph and global-feature encoders joined by late concatenation, temperature-scaled Monte Carlo dropout with conformal-style prediction sets, applicability-domain analysis, and SHAP-guided toxicophore discovery with occlusion controls. On the leakage-controlled test fold, a five-seed soft-voting ensemble achieved a Matthews correlation coefficient (MCC) score of 0.44, an area under the receiver operating characteristic curve (AUROC) score of 0.83, and an area under the precision-recall curve (AUPRC) score of 0.58. It exceeded four ECFP4-based shallow baselines on all 11 endpoints under the same split and validation-based threshold-selection protocol. Controlled ablations showed that the global pathway was important, whereas late concatenation outperformed the tested gated and feature-wise linear modulation fusion variants. Conformal-style prediction sets revealed substantial endpoint-specific variation in set efficiency, and discrimination and calibration improved with similarity to the training domain. Retraining on fixed published Tox21 Challenge and Therapeutics Data Commons folds produced competitive, but not uniformly state-of-the-art, performance. SHAP-guided occlusion and consensus subgraph mining yielded model-derived structural hypotheses, 44 of which contained at least one occurrence that passed the predefined counterfactual criteria.

**Scientific Contribution.** ToxLens integrates leakage-controlled partitioning, uncertainty reporting, applicability-domain auditing, and counterfactually controlled structural interpretation within one reproducible multi-task toxicity workflow. Its scientific contribution is the auditable combination and fixed-split evaluation of these components, rather than a claim of universal state-of-the-art discrimination.

# 1. Introduction

Chemical toxicity is the capacity of a substance to cause adverse effects in biological systems. Computational toxicity prediction has therefore become an important component of early compound triage [11, 16, 17]. Experimental toxicity assays are costly and time-consuming, and they cannot screen chemical libraries at the scale required for modern drug discovery, chemical-safety assessment, and environmental prioritisation [18-23]. Machine-learning models can complement these assays by estimating toxicity risk from molecular structure, thereby supporting compound deprioritisation, redesign, or selection for confirmatory testing.

Despite rapid progress in molecular machine learning, practical reliability cannot be inferred from benchmark metrics alone [8, 105]. Reported performance can be inflated when structurally similar molecules, near-duplicates, or closely related chemotypes occur across training and test folds [24-28, 33]. In such settings, a model may appear to generalise while relying partly on interpolation among close analogues [31, 52]. This issue is especially important in prospective toxicity prediction, where query compounds may be structurally distinct from the training distribution. A useful model should therefore report structural separation, applicability-domain behaviour, and uncertainty alongside predictive discrimination [32, 56, 57].

A second limitation is that many high-capacity molecular models [34–36, 38–40] provide limited chemically interpretable evidence for their predictions [44]. Toxicologists and medicinal chemists often reason in terms of structural alerts, reactive groups, physicochemical liabilities, metabolic activation and known toxicophore patterns [91–93]. A black-box toxicity score, even when accurate on a held-out set, is less useful if it cannot be related back to substructures that plausibly drive the prediction [41]. Post hoc attribution methods can help [15], but saliency maps should not always be treated as evidence on their own [42]. For molecular safety applications, explanations should be tested for faithfulness by perturbing or masking the highlighted substructures and quantifying whether the prediction changes in the expected direction [43].

A third challenge is uncertainty quantification. Binary classifiers are commonly forced to assign every molecule to a toxic or non-toxic class, even when a compound is far from the training distribution or an endpoint is sparse, noisy, or imbalanced [72, 74, 75]. Such forced decisions are poorly matched to chemical triage, where some compounds should instead be flagged for additional testing. Conformal prediction converts model scores into prediction sets with finite-sample marginal coverage under exchangeability [73, 78, 81]. In this setting, a singleton set represents a precise class assignment, whereas a two-label set explicitly communicates that the calibrated evidence is insufficient to distinguish the classes [79, 80].

To address these challenges, we developed ToxLens, a reproducible, leakage-aware framework for multi-task molecular toxicity prediction. ToxLens combines chemical standardisation and conservative duplicate-label handling [29, 30, 37], sphere-exclusion filtering [47], a UMAP-HDBSCAN molecular partition designed to reduce structural leakage [4, 53, 54], a graph neural network paired with a global molecular-feature pathway [1, 2], temperature-scaled Monte Carlo dropout with conformal-style prediction sets [72, 73], and a SHAP-guided toxicophore-mining workflow [15]. The model is evaluated on 11 endpoints comprising Ames mutagenicity [5], acute oral toxicity [6], hERG inhibition [7], and Tox21 nuclear-receptor and stress-response assays [10]. On the leakage-controlled split, the five-seed ensemble achieved macro MCC 0.44, macro AUROC 0.83, and macro AUPRC 0.58, and exceeded ECFP4-based shallow baselines [9] on all 11 endpoints. Separate architecture-level benchmarks retrained ToxLens on fixed published Tox21 Challenge and Therapeutics Data Commons folds to assess transfer under externally defined data partitions.

ToxLens demonstrates how molecular toxicity models can be evaluated and deployed with a broader reliability stack: leakage-controlled validation, applicability-domain auditing, calibrated abstention, reproducible external benchmarking and chemically grounded interpretation. This combination is intended to make toxicity prediction more transparent, reproducible and useful for compound prioritisation.

# 2. Methods

The ToxLens workflow comprises endpoint acquisition and curation (Section 2.1), chemical standardisation and partitioning (Section 2.2), feature engineering (Section 2.3), model architecture and training (Section 2.4), post-hoc uncertainty quantification (Section 2.5), ablation analysis (Section 2.6), interpretation (Sections 2.7-2.8), and endpoint reportability auditing (Section 2.9). This workflow preserves strict separation of training, validation, and test uses.

## 2.1. Data Acquisition and Curation

The curation pipeline enforces three properties: resolution of duplicate-label conflicts, masking of missing labels under the multi-task loss, and removal of near-duplicate structures prior to splitting (Section 2.2).

To mitigate endpoint-specific data sparsity, we used hard parameter sharing in a multi-task learning framework. A shared graph encoder and shared global-feature encoder learn representations used by all task heads, while group-specific towers and endpoint-specific output layers permit task-level specialisation. This design follows the multi-task learning rationale that related tasks can provide useful inductive bias through shared representations [48, 49]. The endpoints differ biologically, but several involve overlapping physicochemical drivers and molecular initiating events [50]. Multi-task sharing can therefore improve data efficiency, although negative transfer remains possible and is assessed through endpoint-level results rather than assumed from the architecture.

The model is an 11-task binary classifier covering bacterial mutagenicity (Ames) [5], acute oral toxicity (LD50_Zhu) [6], hERG potassium-channel inhibition (hERG_Karim) [7], aryl hydrocarbon receptor activity (NR-AhR), aromatase activity (NR-Aromatase), oestrogen receptor activity (NR-ER), oestrogen receptor ligand-binding-domain activity (NR-ER-LBD), antioxidant response element signalling (SR-ARE), heat-shock response (SR-HSE), mitochondrial membrane-potential disruption (SR-MMP), and p53 stress-response activity (SR-p53). The eight nuclear-receptor and stress-response endpoints were obtained from Tox21 [10]. Continuous LD50_Zhu replicates were averaged by canonical SMILES before binarisation at the configured threshold of 2.5. For endpoints already supplied as binary labels, canonical-SMILES duplicates were retained only when their available labels agreed; conflicting labels were set to missing and excluded from the masked loss for that endpoint.

Two external architecture-level evaluations were performed. First, the 12-endpoint Tox21 Challenge archive [10, 11] was used with its canonical training and final-test partitions. Second, single-task TDC ADMET benchmark folds were used for AMES, DILI, and hERG comparisons with TDC leaderboard reference values associated with the published Chemprop, CMPNN, and AttentiveFP architectures [12-14]. For each external benchmark, the benchmark test fold remained unchanged, a 90:10 training/validation split was created only within the development fold, and ToxLens was retrained from scratch. These experiments evaluate the architecture and featurisation under externally specified splits; they are not external validation of the fitted 11-endpoint in-house checkpoint.

## 2.2. Chemical Standardisation and Chemistry-Aware Data Partitioning

Chemical structures were standardised with RDKit [51]. The pipeline selected the largest fragment, applied charge neutralisation and tautomer canonicalisation, sanitised the molecule, generated a canonical isomeric SMILES representation [37], and added explicit hydrogen atoms for graph construction. Standardisation preceded duplicate resolution and split assignment so that equivalent structures were compared in a common representation.

For binary endpoints, duplicate molecules were resolved by strict consensus. When replicate records with the same canonical SMILES contained both class labels, the endpoint label for that structure was set to missing rather than assigned by majority vote. For LD50_Zhu, which was supplied as a continuous measurement, replicate values were first averaged and then binarised. This distinction avoids treating a continuous replicate measurement as if it were a conflicting binary annotation. The curation follows QSAR best-practice recommendations concerning salts, duplicates, ambiguous labels, and reproducible descriptor generation [29-31].

Before constructing the selected UMAP-HDBSCAN split, sphere exclusion was applied in Morgan-fingerprint space using RDKit LeaderPicker [47]. LeaderPicker operates on distance, so a Tanimoto-similarity ceiling of 0.95 was implemented as a distance threshold of 0.05. The retained representatives were then used for the UMAP-HDBSCAN partition; the procedure was not retroactively applied to the random, Butina, or scaffold comparator splits.

The retained molecules were divided into training, validation, and held-out test subsets using a UMAP-HDBSCAN strategy adapted from structured molecular-splitting work [4]. Radius-2, 1,024-bit Morgan fingerprints [9] were projected into 10 dimensions with UMAP [45, 53] using Jaccard distance (100 neighbours, minimum distance 0.15, seed 42),

L2-normalised, and clustered with HDBSCAN [46, 54] (minimum cluster size 15; minimum samples 5). HDBSCAN noise points were treated as singleton clusters. Clusters, rather than individual molecules, were assigned to 80:10:10 folds. Stratification used an any-positive proxy equal to one when at least one available primary endpoint was positive for a molecule; the cluster-level proxy was the mean of this indicator, dichotomised at the median cluster mean. This preserves cluster separation while reducing severe fold-level label imbalance. Random, Butina [55], Bemis-Murcko scaffold, and UMAP-HDBSCAN splits were compared using structural novelty, scaffold novelty, property balance, and aggregate out-of-distribution utility (Supplementary Table 3 and Supplementary Figure 1). These diagnostics characterise a trade-off for this dataset and do not establish UMAP-HDBSCAN as a universally optimal split.

The sphere-exclusion and UMAP-HDBSCAN procedures were used only for the in-house 11-endpoint panel. External benchmarks retained their published development/test partitions. Validation subsets were created exclusively from each benchmark's development fold, and the published test fold was not used for model selection, threshold fitting, probability calibration, or conformal calibration.

## 2.3. Feature Engineering

Each molecule was represented as an explicit-hydrogen graph with 134 node features and 35 edge features [58-64]. Node features encode element, degree, formal charge, chirality, hydrogen count, hybridisation, aromaticity, atomic mass, Gasteiger charge, ring membership, atom-level pharmacophore flags, Crippen and topological polar surface area contributions, electrotopological state and Labute surface-area contributions, electronegativity-derived quantities, valence and neighbourhood descriptors, and SMARTS-derived atom-reactivity flags. Edge features encode bond order, stereochemistry, conjugation, ring membership, electronegativity and charge differences, ring size, rotatability, heteroatom and halogen context, and SMARTS-derived bond-reactivity flags. The complete ordered feature definitions are supplied in the reproducibility materials.

The stored global-feature tensor had width 3,190 and concatenated six blocks in a fixed order: a 1,024-bit radius-2 Morgan fingerprint [9], 217 RDKit descriptors [51], 76 toxicity-pattern features (75 SMARTS matches plus one PAINS count) [63, 65], a 768-dimensional MolFormer-XL embedding [38], 905 conformer-derived RDKit three-dimensional descriptors, and a reserved 200-dimensional PubChem-bioactivity block. The three-dimensional block comprised WHIM, GETAWAY, USR, USRCAT, MORSE, RDF, and 12 scalar shape descriptors from a deterministically seeded, MMFF-optimised conformer. The PubChem cache was absent for the reported experiments; consequently, all 200 PubChem entries were zero and contributed no predictive information. The effective non-zero feature width was therefore 2,990, although the checkpoint input schema remained 3,190 dimensions.

## 2.4. Neural Network Architecture

The model contains parallel graph and global-feature encoders. The graph trunk linearly projects the 134 node features to a 256-dimensional hidden representation and applies five residual GINE message-passing layers [2, 68]. Each layer uses graph normalisation, a residual GINE update, stochastic depth, and a residual feed-forward block. A virtual-node state is updated from pooled graph information and injected into subsequent layers to propagate graph-level context. The retained checkpoint used edge-drop probability 0.06, maximum stochastic-depth probability 0.20, and hidden-state dropout 0.20.

After the final message-passing layer, multi-head attention pooling is combined with mean and max pooling to obtain a graph embedding. In parallel, the global vector is sanitised, input-dropped, normalised, and processed by an independent feed-forward projection. The retained model uses no global-to-node modulation, no late residual gate, and no mid-trunk concatenation. Instead, the graph and global representations are concatenated only at the prediction head. This late-concatenation design was selected because the controlled ablation showed better test MCC than the tested GCMI and FiLM variants (Section 4.3).

The concatenated representation is passed through a shared feed-forward trunk and task-specific prediction heads. Task names are grouped into broad toxicity families (genotoxicity, stress response, nuclear receptor, cardio/systemic, or general toxicity), and group-specific towers are added to the shared representation before each task head, implementing hard parameter sharing while still permitting related endpoint groups to learn task-family-specific transformations. The output is one logit per endpoint.

Five models with random seeds 42-46 were trained under the same fixed configuration. For each seed, the checkpoint with the highest validation MCC was retained. The retained checkpoint produced test logits, which were converted to probabilities with the sigmoid function. The final ensemble probability for each molecule and endpoint was the arithmetic mean of the five checkpoint probabilities; task-specific thresholds were selected from the averaged validation probabilities and then frozen for test scoring. This is a soft-voting ensemble and does not assume independent errors among the component models.

## 2.5. Uncertainty Quantification via MC Dropout and Conformal-Style Prediction Sets

Monte Carlo (MC) dropout provides a stochastic estimate of predictive variability but does not itself guarantee frequentist coverage [72, 74, 75]. Split conformal prediction provides finite-sample marginal coverage under exchangeability [73, 78]. ToxLens combines these procedures post hoc: a single temperature parameter is fitted on validation logits by minimising binary negative log-likelihood [104], 30 dropout-enabled forward passes are used to estimate the temperature-scaled predictive mean and standard deviation, and the predictive mean is conformalised separately for each endpoint. The standard deviation is reported as a supplementary uncertainty measure; it is not used to construct the conformal set.

For endpoint t and validation example i, with predictive mean probability p_it and binary label y_it, the nonconformity score is 1 - p_it when y_it = 1 and p_it when y_it = 0. Missing endpoint labels are excluded. For n valid calibration examples, the endpoint quantile is the higher empirical quantile at min(1, ceil((n + 1)(1 - alpha))/n), with alpha = 0.05. At inference, class 0 is included when p_t is no greater than the calibrated quantile, and class 1 is included when 1 - p_t is no greater than that quantile. If neither class is included, the implementation returns the probability argmax; if both are included, the output is flagged as uncertain. Calibration uses the validation fold only.

Standard split-conformal coverage is marginal rather than molecule-specific or class-conditional and assumes that a fixed predictor is evaluated on an exchangeable, independently held-out calibration set [73, 78, 81]. Here, the validation fold was also used for checkpoint selection and temperature fitting; consequently, the nominal finite-sample coverage guarantee does not strictly apply. We therefore treat $\alpha = 0.05$ as a target operating level and report singleton versus two-label-set rates descriptively. MC-dropout variability and applicability-domain similarity remain separate diagnostics.

## 2.6. Ablation Studies and Diagnostic Analyses

Two ablation grids were used to isolate major design choices. The architecture grid (A1-A4) compared full GCMI, removal of fusion, removal of the global pathway, and replacement of the deep head with a linear head. The fusion grid (F1-F4) compared GCMI, FiLM [3], mid-trunk concatenation, and no mid-trunk fusion under a fixed trunk and head. These variants were selected to test three specific questions: whether global features add information beyond the graph, whether cross-modal modulation improves on late concatenation, and whether prediction-head depth materially affects performance. Every configuration was trained from scratch on the same training fold, selected by validation MCC, and evaluated on the same held-out test fold. MCC, AUROC, AUPRC, and Brier score were reported; AUPRC and proper scoring rules are included because several endpoints are imbalanced and AUROC alone can obscure poor positive-class precision [82-85].

## 2.7. Interpretability

The interpretability analysis comprises atom-level GradientSHAP attribution, a quantitative SHAP-guided occlusion audit, and dataset-wide consensus motif mining (Section 2.8). All analyses use the same retained single-model checkpoint as the headline single-model test results. Machine-readable per-molecule attribution, faithfulness, and toxicophore tables accompany the rendered figures.

Applying Captum attribution to a PyTorch Geometric model requires preserving graph topology while Captum perturbs attributed tensors [86, 88]. We therefore used PyG-Captum-SHAP version 0.1.5, which passes node, edge, and global-feature tensors as attributed inputs while shielding and reconstructing the block-diagonal edge-index structure during sampled forward passes. GradientSHAP [15, 88] was applied jointly to the three input modalities. Because attribution maps alone do not demonstrate causal or mechanistic faithfulness [42, 43, 89, 90], the manuscript interprets them only together with graph-feature occlusion controls and consensus-motif statistics.

For visualisation, explicit-hydrogen attributions were collapsed onto heavy atoms. Each hydrogen-node attribution and each heavy-hydrogen edge attribution was added to its parent heavy atom, whereas heavy-heavy edge attributions remained associated with the corresponding undirected bond. This aggregation preserves the total summed attribution assigned to the displayed atom-bond system while producing conventional heavy-atom depictions. It is a rendering transformation, not a second model evaluation. Similarity-map approaches provide related precedents for projecting model contributions onto chemically interpretable atom environments [64].

## 2.8. Automated Toxicophore Discovery Algorithm

To move from molecule-specific attributions to recurring structural hypotheses, positive GradientSHAP atom scores were mined across positive test examples. After hydrogen collapse, each heavy-atom score vector was divided by its maximum absolute value. Candidate seed atoms had to exceed 0.20 and the 90th percentile of strictly positive attributions within the molecule; at most three seeds were retained. Each seed was expanded over one or two bonds, stopping at the smallest radius that captured at least 35% of the molecule's total positive attribution mass. Resulting fragments were canonicalised as SMILES after the same molecule-level standardisation used for modelling. These thresholds were prespecified in the analysis script and should be interpreted as algorithmic design choices rather than chemically validated boundaries.

Recurring fragments were grouped using 1,024-bit radius-2 Morgan fingerprints [9] and DBSCAN [94] with Jaccard distance. Clusters required at least three occurrences. Exact-repeat grouping was used as a fallback when density clustering did not return a cluster. This procedure groups similar model-attributed fragments; it does not establish that every member shares a biological mechanism.

Consensus-fragment occurrences were evaluated by graph-feature occlusion. For a positive test molecule, node features for the selected heavy atoms and attached hydrogens, together with incident edge features, were set to zero while topology and global features were held fixed. The change in predicted positive-class probability was compared with up to 20 connected, size-matched random subgraphs from the same molecule. An occurrence passed the predefined counterfactual criteria when the probability drop was at least 0.10, exceeded the mean random-control drop by at least 0.05, and had a one-sided empirical p-value no greater than 0.10. This is a within-model feature-occlusion test, not chemical deletion, causal intervention on a real molecule, or experimental validation.

Clusters were ranked by the fraction of tested occurrences passing the counterfactual criteria, mean probability drop, difference from random controls, and prevalence among positive molecules. A cluster was labelled counterfactually supported when at least one tested occurrence passed all three criteria; clusters with no passing occurrence were retained only as SHAP-consensus results. The reported motifs are therefore model-derived structural hypotheses. Their chemical interpretation requires the representative parent molecules, occurrence counts, and counterfactual statistics, and should not be inferred from short fragment SMILES alone.

## 2.9. Test-Set Reportability Gate

Endpoint-level estimates were screened with a prespecified reportability gate before inclusion in external-benchmark headline macro means: total test count N had to be at least 100 and the minority-class count had to be at least 15. Endpoints failing the gate were still reported individually but were excluded from the gated macro summary. All 11 in-house endpoints passed the gate. On the Tox21 Challenge final test set, NR-AR (12 positives) and NR-AR-LBD (8 positives) failed the minority-class criterion, leaving 10 endpoints in the gated macro. This threshold is a reporting convention for this study, not a universal statistical standard; ungated 12-task Tox21 results are also reported for comparability with published work.

## 3. Software and Reproducibility Resources

The reproducibility package supplied with this submission contains the source code, a readable Conda environment specification and an exact Windows package lock, fixed train/validation/test assignments, selected single-model and ensemble checkpoints, benchmark-provenance records, and machine-readable result artefacts. PyG-Captum-SHAP version 0.1.5, used by the interpretability workflow, is available from the Python Package Index. No public prediction server is required to reproduce the reported analyses.

## 4. Results

### 4.1. Per-Task Performance with Confidence Intervals

Single-model performance was evaluated using the checkpoint with the highest validation MCC. Task-specific thresholds were selected on the validation fold and frozen before test scoring. Table 1 reports 1,000-resample percentile bootstrap 95% confidence intervals for the held-out test fold; Section 4.2 reports the separately trained five-seed soft-voting ensemble used for the primary aggregate comparison.

**Table 1: Per-task Test-Set Performance with 95% Bootstrap Confidence Intervals (CI).** Decision thresholds are validation-optimised under the per-task MCC objective. Bounds are 1,000-iteration percentile bootstrap CIs at the per-task validation-optimal probability threshold. The macro row reports the unweighted mean across endpoints.

| Endpoint | Evaluation Metrics | | | |
|---|---|---|---|---|
| | MCC [95 % CI] | AUROC [95 % CI] | AUPRC [95 % CI] | Brier [95 % CI] |
| Ames | 0.54 [0.49, 0.60] | 0.86 [0.83, 0.88] | 0.90 [0.88, 0.93] | 0.14 [0.13, 0.16] |
| LD50_Zhu | 0.42 [0.35, 0.49] | 0.79 [0.75, 0.82] | 0.69 [0.64, 0.74] | 0.19 [0.17, 0.21] |
| hERG_Karim | 0.39 [0.34, 0.44] | 0.75 [0.73, 0.78] | 0.74 [0.71, 0.78] | 0.22 [0.20, 0.23] |
| NR-AhR | 0.45 [0.35, 0.55] | 0.86 [0.81, 0.90] | 0.62 [0.52, 0.71] | 0.09 [0.07, 0.10] |
| NR-Aromatase | 0.26 [0.08, 0.45] | 0.81 [0.73, 0.87] | 0.29 [0.16, 0.46] | 0.07 [0.05, 0.08] |
| NR-ER | 0.35 [0.23, 0.48] | 0.77 [0.70, 0.85] | 0.46 [0.35, 0.59] | 0.08 [0.06, 0.09] |
| NR-ER-LBD | 0.55 [0.38, 0.70] | 0.88 [0.80, 0.94] | 0.59 [0.43, 0.75] | 0.03 [0.02, 0.03] |
| SR-ARE | 0.29 [0.20, 0.38] | 0.77 [0.73, 0.82] | 0.51 [0.43, 0.59] | 0.16 [0.14, 0.17] |
| SR-HSE | 0.22 [0.09, 0.36] | 0.82 [0.73, 0.89] | 0.21 [0.12, 0.39] | 0.05 [0.04, 0.05] |

| Endpoint | Evaluation Metrics | | | |
|---|---|---|---|---|
| | MCC [95 % CI] | AUROC [95 % CI] | AUPRC [95 % CI] | Brier [95 % CI] |
| SR-MMP | 0.62 [0.55, 0.69] | 0.90 [0.87, 0.93] | 0.79 [0.73, 0.84] | 0.14 [0.12, 0.16] |
| SR-p53 | 0.34 [0.22, 0.46] | 0.84 [0.79, 0.89] | 0.32 [0.23, 0.46] | 0.06 [0.05, 0.07] |
| **Macro** | 0.40 | 0.82 | 0.56 | 0.11 |

The retained single checkpoint achieved macro test MCC 0.40, macro AUROC 0.82, macro AUPRC 0.56, and macro Brier score 0.11 across the 11 endpoints. All AUROC point estimates were at least 0.75, and seven exceeded 0.80. The highest single-model AUROC values occurred for SR-MMP (0.90), NR-ER-LBD (0.88), Ames (0.86), NR-AhR (0.86), and SR-p53 (0.84), whereas hERG_Karim was lowest (0.75). Threshold-dependent MCC was more heterogeneous, ranging from 0.22 for SR-HSE to 0.62 for SR-MMP. This divergence illustrates why ranking, thresholded classification, and probability-quality metrics are reported together rather than treating any one metric as sufficient. Per-task BEDROC, enrichment factors, and Cohen's kappa are reported in Supplementary Table 6. The corresponding per-task ROC curves are shown in Figure 1.

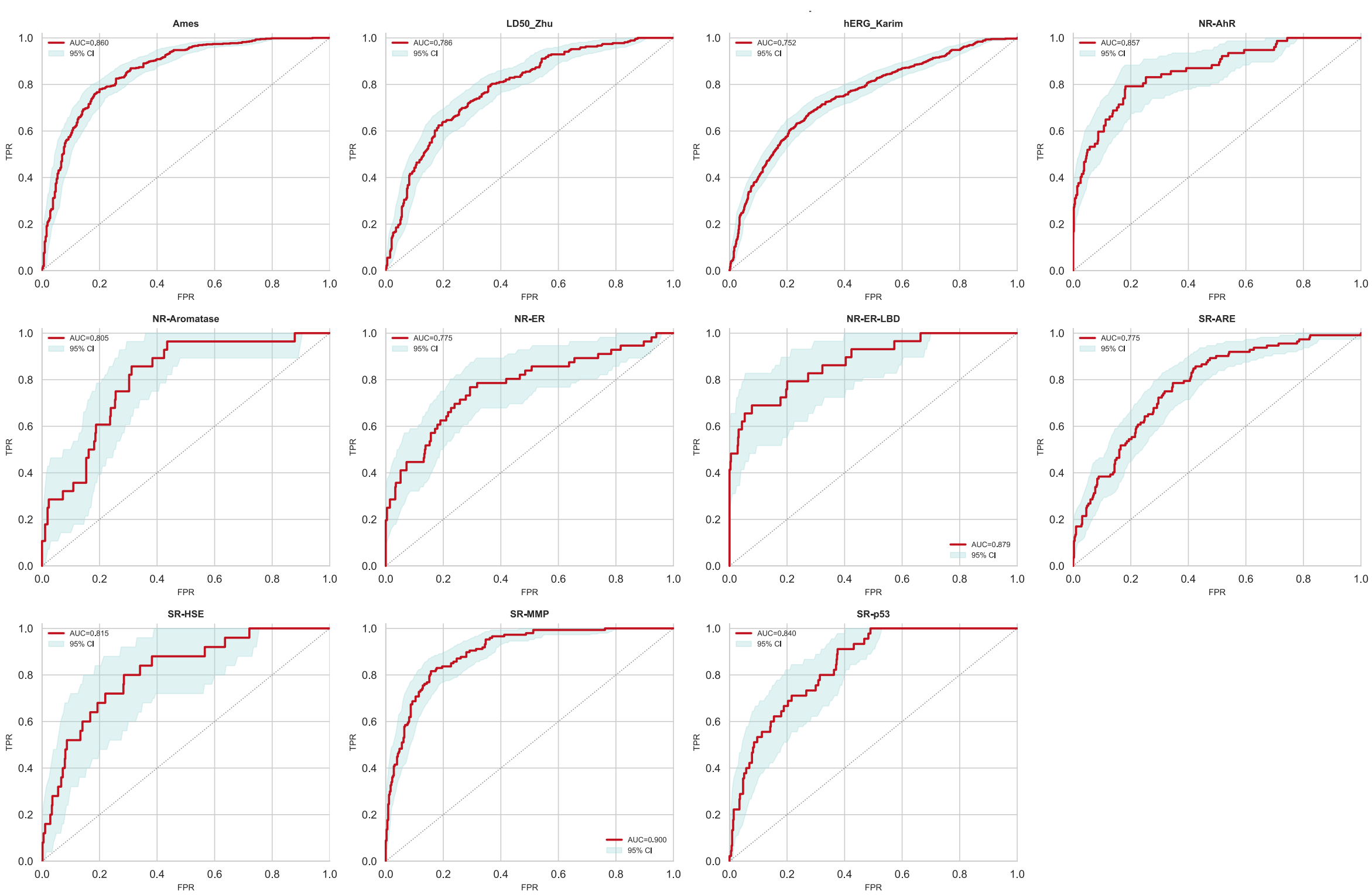

**Figure 1: Per-task receiver operating characteristic curves for the retained single-model checkpoint.** Each panel shows one primary endpoint, the no-skill diagonal, and a pointwise 95% percentile-bootstrap band based on 1,000 test-fold resamples. Table 1 gives the corresponding AUROC point estimates and confidence intervals. Band width depends on endpoint sample size, class balance, and score distribution; it should not be interpreted as evidence of mechanistic certainty.

## 4.2. Five-Seed Ensemble Aggregate and Shallow-Baseline Comparison

The five-seed soft-voting ensemble described in Section 2.4 is the primary aggregate for the in-house UMAP-HDBSCAN evaluation. Four shallow comparators - Random Forest [96], XGBoost [97], a multilayer perceptron [98], and a support-vector machine [99] - were trained separately for every endpoint on the same 1,024-bit ECFP4 representation [9] and the identical training, validation, and test folds. For every model, the task-specific decision threshold was selected on validation probabilities by maximising MCC and then frozen for test scoring.

Across the 11 endpoints, the ToxLens ensemble achieved macro test MCC 0.44, macro AUROC 0.83, and macro AUPRC 0.58. Corresponding shallow-baseline macro MCC and AUROC values were 0.32 and 0.78 for Random Forest, 0.29 and 0.74 for XGBoost, 0.25 and 0.69 for the multilayer perceptron, and 0.20 and 0.66 for the support-vector machine (Supplementary Table 2). ToxLens had the highest test MCC on all 11 endpoints. The largest absolute margins over the strongest shallow baseline were observed for SR-MMP (+0.19), SR-HSE (+0.16), NR-ER (+0.15), and SR-p53 (+0.13); the smallest were observed for NR-AhR (+0.03), Ames (+0.04), and hERG_Karim (+0.07). These are descriptive endpoint-level comparisons; no formal superiority claim is made because the 11 endpoints are heterogeneous and only one fixed split was evaluated. The per-endpoint MCC comparison is shown in Figure 2.

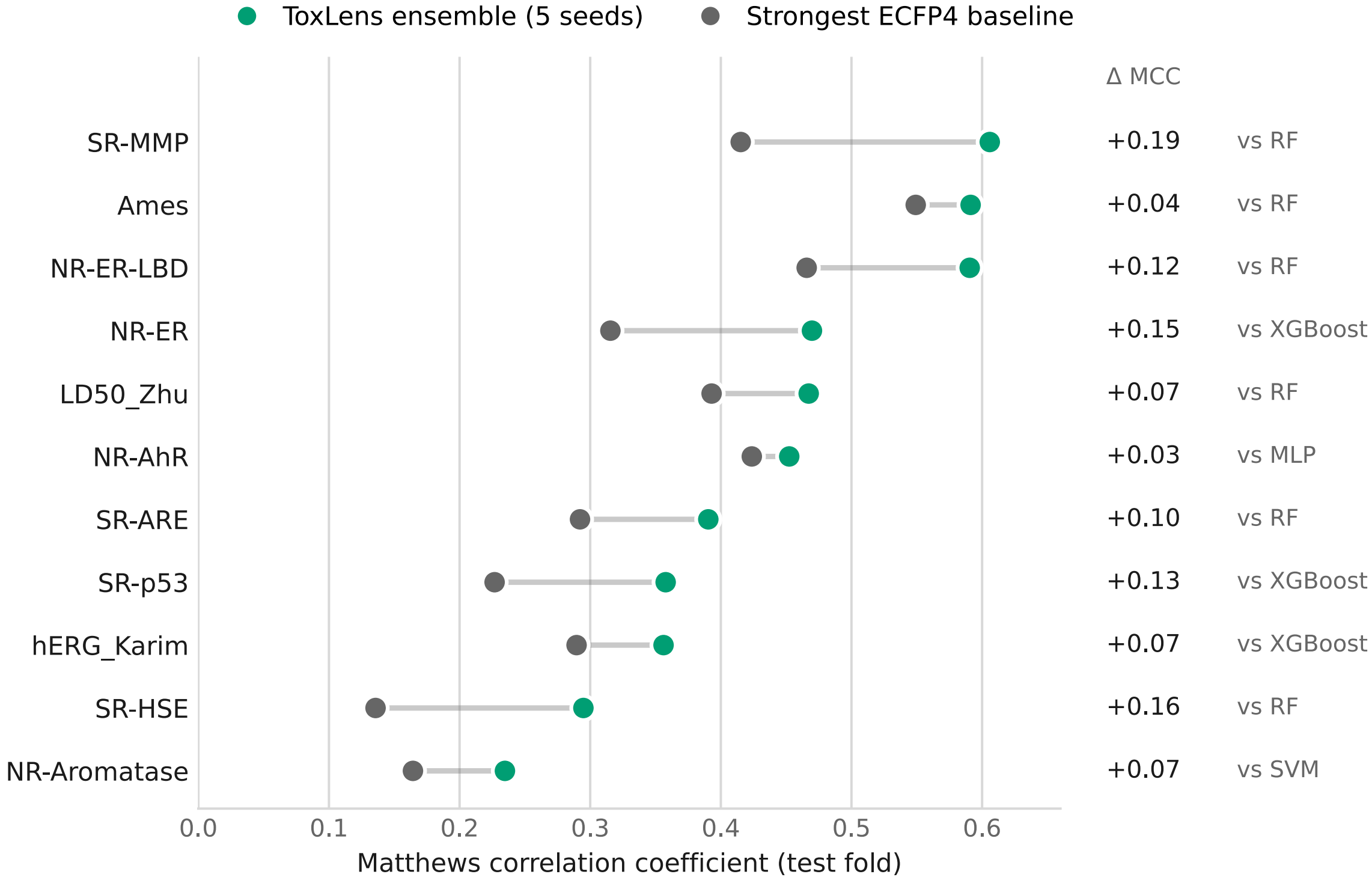

**Figure 2: Per-endpoint held-out test MCC for the five-seed ToxLens ensemble and the strongest ECFP4 shallow baseline.** For each endpoint, the strongest shallow comparator is selected from Random Forest, XGBoost, multilayer perceptron, and support-vector machine. Every model used the same split and a threshold selected on its own validation probabilities. The plot is descriptive: ToxLens achieved the highest MCC for all 11 endpoints, while the identity of the strongest shallow comparator varied by endpoint. Complete values for all models are reported in Supplementary Table 2.

Relative to the retained single checkpoint, ensembling increased macro MCC from 0.40 to 0.44 and macro AUROC from 0.82 to 0.83. MCC increased on eight endpoints and decreased on hERG_Karim (-0.04), NR-Aromatase (-0.03), and SR-MMP (-0.02). Because thresholds were selected separately from averaged validation probabilities, these endpoint-level changes reflect both probability averaging and movement of the validation-selected operating point. They should not be described as an intrinsic precision-recall trade-off without a prespecified operating-cost analysis.

## 4.3. Architectural Ablation

Eight controlled variants were trained under the same split and model-selection protocol. The architecture grid comprised A1 (full GCMI), A2 (no mid-trunk fusion with global features), A3 (GNN only), and A4 (GCMI with a linear head). The fusion grid comprised F1 (GCMI), F2 (FiLM), F3 (mid-trunk concatenation), and F4 (no mid-trunk fusion). Table 2 reports validation and held-out test results for each independently trained variant.

Three findings emerged from the ablation grid. First, removing the global pathway produced the weakest tested configuration (A3: test MCC 0.341, AUROC 0.791), supporting the contribution of global features beyond the graph representation. Second, the two no-mid-trunk-fusion runs produced the highest test MCC values (A2: 0.438; F4: 0.425), whereas the GCMI runs achieved 0.382 and 0.373. FiLM and mid-trunk concatenation were intermediate at 0.398 and 0.403. Third, replacing the deep head with a linear head within the GCMI architecture increased MCC from 0.382 to 0.407, indicating that head depth was not the principal determinant of performance in this grid. These are single-run ablations and therefore quantify observed differences, not uncertainty over retraining.

The ablation supports late concatenation as the default for this endpoint panel. One plausible explanation is that graph and global encoders carry complementary information that is degraded when global features modulate node channels during message passing. That mechanism was not directly tested, however, and should be treated as a hypothesis rather than a demonstrated cause. The controlled conclusion is narrower: under the fixed split and training protocol, the tested GCMI and FiLM variants did not improve on the simpler no-mid-trunk-fusion configuration. The formal GCMI definition is retained in Supplementary Section 2.3 solely to document the comparator.

**Table 2: Architecture and Fusion Ablation on the UMAP-HDBSCAN Held-Out Test Fold.** Each row is one independently trained run selected by validation MCC. A2 and F4 implement nominally equivalent no-mid-trunk-fusion architectures but arose from separate training runs and therefore quantify run-to-run variation rather than a deterministic configuration difference.

| Configuration | Fusion | Global | Head | AUROC | MCC | AUPRC | Brier |
|---|---|---|---|---|---|---|---|
| A2 (No-Fusion, deep) | None | ✓ | deep | 0.805 | **0.438** | 0.558 | 0.114 |
| F4 (No-Fusion, deep) | None | ✓ | deep | 0.827 | 0.425 | 0.564 | 0.103 |
| A4 (GCMI + linear head) | GCMI | ✓ | linear | 0.815 | 0.407 | 0.556 | 0.110 |
| F3 (Concat) | Concat | ✓ | deep | 0.828 | 0.403 | 0.561 | 0.109 |
| F2 (FiLM) | FiLM | ✓ | deep | 0.817 | 0.398 | 0.546 | 0.107 |
| A1 (Full GCMI) | GCMI | ✓ | deep | 0.815 | 0.382 | 0.548 | 0.110 |
| F1 (GCMI) | GCMI | ✓ | deep | 0.820 | 0.373 | 0.565 | 0.106 |
| A3 (GNN-only) | None | ✗ | deep | 0.791 | 0.341 | 0.517 | 0.114 |

## 4.4. Conformal-Style Prediction-Set Efficiency

Conformal-style prediction sets were evaluated at $\alpha = 0.05$ using the validation fold for calibration. The proportion of singleton prediction sets varied markedly by endpoint (Table 3). Singleton efficiency exceeded 80% for NR-AhR, NR-Aromatase, NR-ER-LBD, SR-HSE, and SR-p53, and reached 100% for NR-ER-LBD. It was below 50% for Ames, LD50_Zhu, hERG_Karim, and SR-ARE. A two-label set indicates that both classes satisfy the endpoint-specific calibration threshold and is therefore flagged for further review. These percentages describe prediction-set efficiency, not per-molecule correctness. Because the same validation fold informed checkpoint selection, temperature scaling, and calibration, $\alpha = 0.05$ is a target operating level rather than a formal 95% coverage guarantee.

**Table 3: Conformal-Style Prediction-Set Efficiency. Calibration used α = 0.05.** Precise denotes a singleton set, and uncertain denotes the two-label set {0, 1}. Percentages measure set efficiency, not singleton accuracy or formal coverage.

| Task Endpoint | Precise Predictions (%) | Uncertain / Flagged (%) |
|---|---|---|
| Ames | 42.39 | 57.61 |
| LD50_Zhu | 20.40 | 79.60 |
| hERG_Karim | 47.51 | 52.49 |
| NR-AhR | 87.58 | 12.42 |
| NR-Aromatase | 93.30 | 6.70 |
| NR-ER | 63.80 | 36.20 |
| NR-ER-LBD | 100.00 | 0.00 |
| SR-ARE | 34.65 | 65.35 |
| SR-HSE | 89.83 | 10.17 |
| SR-MMP | 71.01 | 28.99 |
| SR-p53 | 83.50 | 16.50 |

## 4.5. Applicability-Domain Stratified Performance

Test molecules were binned into quartiles by maximum Tanimoto similarity to the training set, with Q1 the most distant and Q4 the most similar. Mean AUROC increased from 0.77 in Q1 to 0.86 in Q4, mean MCC from 0.28 to 0.44, and mean expected calibration error decreased from 0.13 to 0.08 (Supplementary Table 4). These aggregate trends are consistent with better performance near the training domain, although individual endpoints were not uniformly monotonic.

Endpoint-level behaviour was heterogeneous. NR-ER-LBD showed a strong increase in AUROC across similarity quartiles (0.71, 0.81, 0.95, and 0.96), whereas Ames varied only modestly. Several endpoints departed from monotonicity because quartile-specific positive counts were small. The applicability-domain analysis should therefore be interpreted as a panel-level association between structural proximity and performance, not as a deterministic reliability rule for each molecule. Similarity quartile, MC-dropout variability, and conformal set size provide complementary diagnostics.

## 4.6. External Validation Benchmarks

External experiments evaluated the ToxLens architecture under externally defined benchmark partitions rather than applying the fitted in-house checkpoint to unseen endpoints. For every benchmark, the development and test folds were preserved; a validation subset was created only from the development fold; featurisation and model fitting were repeated from scratch; checkpoint and threshold selection used validation data; and the test fold was scored only after selection. Canonical-SMILES overlap between development and test folds was audited. TDC comparisons use leaderboard reference values associated with the same benchmark split; the comparator models were not rerun in this study.

#### 4.6.1. Tox21 Challenge

ToxLens was retrained on the canonical Tox21 Challenge development partition and evaluated on the unchanged final-test partition [10, 11]. Temperature scaling, task-specific thresholds, and conformal calibration used a validation slice drawn only from the challenge development set. Across all 12 endpoints, macro AUROC was 0.82, macro AUPRC 0.35, macro MCC 0.34, and macro Brier score 0.06. NR-AR (12 positives) and NR-AR-LBD (8 positives) failed the prespecified minority-class reportability gate and were excluded from the gated headline, leaving 10 endpoints with macro AUROC 0.82, macro AUPRC 0.38, macro MCC 0.37, and macro Brier score 0.07. The ungated 12-task result remains the appropriate aggregate for direct comparison with published 12-task Tox21 means. Per-endpoint test metrics are reported in Table 4, with extended enrichment and BEDROC results in Supplementary Table 5.

Discrimination was strongest for SR-MMP (AUROC 0.95), NR-AhR (0.91), and NR-AR (0.85; excluded from the gated macro). AUROC was lowest for NR-PPAR-gamma (0.75) and NR-ER-LBD (0.77). Confidence intervals were wide for endpoints with few positives, and thresholded MCC was correspondingly unstable. Top-ranked enrichment remained high for several sparse endpoints, but enrichment factors are sensitive to prevalence and shortlist size and are reported as triage-oriented ranking diagnostics rather than substitutes for AUROC, AUPRC, or MCC.

**Table 4: ToxLens Performance on the Tox21 Challenge Final Test Set.** Thresholds were selected on the challenge validation subset; intervals are 1,000-resample percentile-bootstrap 95% confidence intervals. Dagger-marked endpoints fail the prespecified minority-class reportability gate and are excluded only from the gated 10-task macro.

| Endpoint | Positive/ Negative | AUROC | AUPRC | MCC | Brier |
|---|---|---|---|---|---|
| NR-AhR | 71 / 537 | 0.91 [0.87, 0.94] | 0.63 [0.53, 0.74] | 0.56 [0.47, 0.65] | 0.08 |
| NR-Aromatase | 39 / 489 | 0.79 [0.71, 0.86] | 0.24 [0.17, 0.37] | 0.30 [0.18, 0.42] | 0.07 |
| NR-ER | 50 / 465 | 0.79 [0.72, 0.86] | 0.49 [0.37, 0.62] | 0.45 [0.31, 0.58] | 0.07 |
| NR-ER-LBD | 20 / 579 | 0.77 [0.66, 0.86] | 0.23 [0.08, 0.41] | 0.38 [0.00, 0.54] | 0.04 |
| NR-PPAR-gamma | 31 / 572 | 0.75 [0.65, 0.84] | 0.17 [0.11, 0.30] | 0.19 [0.05, 0.33] | 0.05 |
| SR-ARE | 92 / 461 | 0.79 [0.74, 0.84] | 0.44 [0.37, 0.55] | 0.37 [0.29, 0.46] | 0.13 |
| SR-ATAD5 | 37 / 583 | 0.78 [0.70, 0.85] | 0.33 [0.20, 0.48] | 0.37 [0.18, 0.53] | 0.05 |
| SR-HSE | 21 / 587 | 0.81 [0.70, 0.91] | 0.31 [0.17, 0.55] | 0.38 [0.16, 0.59] | 0.05 |
| SR-MMP | 58 / 483 | 0.95 [0.91, 0.97] | 0.63 [0.53, 0.77] | 0.55 [0.43, 0.67] | 0.07 |

| Endpoint | Positive/ Negative | AUROC | AUPRC | MCC | Brier |
|---|---|---|---|---|---|
| SR-p53 | 40 / 574 | 0.84 [0.78, 0.89] | 0.31 [0.21, 0.44] | 0.20 [0.06, 0.34] | 0.07 |
| **10-task macro** | — | **0.82** | **0.38** | **0.37** | **0.07** |
| NR-AR† | 12 / 574 | 0.85 [0.71, 0.97] | 0.37 [0.17, 0.64] | 0.21 [-0.01, 0.46] | 0.02 |
| NR-AR-LBD† | 8 / 573 | 0.79 [0.63, 0.91] | 0.08 [0.03, 0.33] | 0.08 [-0.02, 0.27] | 0.02 |
| **12-task macro (incl. excluded†)** | — | **0.82 [0.73, 0.89]** | **0.35 [0.24, 0.52]** | **0.34 [0.17, 0.49]** | **0.06** |

The 12-task ToxLens mean AUROC of 0.82 was below the published DeepTox DNN-only mean (0.84) and winning ensemble mean (0.85) [11]. Individual endpoint comparisons varied, but they do not establish overall superiority. The principal value of this experiment is protocol reproduction on the canonical challenge partition with modern uncertainty and interpretability outputs, not a state-of-the-art claim.

#### 4.6.2. TDC ADMET Benchmarks

The same retraining protocol was applied to three TDC ADMET classification folds. On AMES, ToxLens achieved mean test AUROC 0.831 ± 0.007 across five development re-splits, exceeding the TDC AttentiveFP reference value (0.814) but trailing CMPNN (0.843) and Chemprop-RDKit (0.850) [12-14]. On DILI, ToxLens achieved 0.887 ± 0.017, close to the TDC AttentiveFP (0.886) and Chemprop-RDKit (0.887) reference values, and below Chemprop (0.899). On hERG, ToxLens achieved 0.805 ± 0.055; this fold is retained as protocol reproduction only because the benchmark development and test partitions contain six canonical duplicate structures. These comparator values are TDC leaderboard entries for the named architectures, not independent reruns performed in this study. Per-seed results, fold sizes, overlap audits, and comparator values are reported in Supplementary Table 7. Aggregate benchmark results are reported in Table 5.

**Table 5: TDC ADMET Retraining on Fixed Benchmark Folds.** ToxLens values are mean ± standard deviation across five 90:10 development re-splits. Every seed was retrained from scratch, selected on validation MCC, and evaluated on the unchanged benchmark test fold. Comparator values are TDC leaderboard point estimates associated with the named published architectures, not reruns by this study. The hERG result is protocol reproduction only because six canonical duplicates occur across its benchmark development and test folds.

| Benchmark | ToxLens | Chemprop | CMPNN | AttentiveFP |
|---|---|---|---|---|
| AMES | 0.83 ± 0.01 | 0.850 | 0.843 | 0.814 |
| DILI | 0.89 ± 0.02 | 0.899 / 0.887 | — | 0.886 |
| hERG† | 0.80 ± 0.05 | 0.840 | — | 0.825 |

Across the fixed-split benchmarks, ToxLens was competitive with the TDC reference architectures but did not uniformly exceed their leaderboard values. The comparisons support transfer of the architecture and featurisation to independently curated folds. They do not show that the fitted 11-endpoint model predicts unseen endpoints, and they do not support a universal state-of-the-art claim. Split design, duplicate structure, class imbalance, and applicability-domain coverage remain major determinants of QSAR benchmark results [27, 28, 31, 33].

## 4.7. Interpretability

The interpretability analysis generated hydrogen-collapsed atom-level attributions, SHAP-guided occlusion results, and a separate consensus-motif catalogue. The saliency and faithfulness analyses used 10 test molecules per endpoint from the retained single-model checkpoint. This small, selected interpretability sample supports a within-model diagnostic analysis, not population-level mechanistic inference. The Tox21 Challenge results are generated by a separately retrained benchmark model and are not part of this interpretability audit.

### 4.7.1. Heavy-Atom SHAP Saliency

Figure 3 illustrates heavy-atom attributions for one molecule under four task heads. Positive and negative signed attributions are normalised within each task-molecule pair, so colour intensity is comparable within a panel but not as an absolute magnitude across endpoints. Overlap among highlighted regions is consistent with the shared encoder, but a single molecule cannot demonstrate cross-task transfer or mechanistic conservation. The figure is therefore illustrative; quantitative support for model reliance on highlighted atoms is provided by the occlusion analysis in Section 4.7.2.

**Figure 3: Heavy-atom GradientSHAP attribution for one molecule across SR-MMP, NR-AhR, NR-ER-LBD, and SR-p53.** Red and blue denote positive and negative signed attribution, respectively, after normalisation by the maximum absolute attribution within each panel; atoms below an absolute normalised attribution of 0.35 are shown in greyscale. The shared highlighted regions are an illustrative observation, not independent evidence of mechanistic transfer.

### 4.7.2. SHAP-Guided Occlusion Faithfulness

At a 0.20 mask fraction, the cross-task mean logit drop was 0.55 for SHAP-guided masking and 0.04 for size-matched random masking, giving a mean faithfulness gap of 0.51 logits (Table 6). The gap was positive for every endpoint and exceeded 0.50 for six endpoints. Median per-molecule empirical p-values were no greater than 0.10 for 10 endpoints; hERG_Karim had a median of 0.16. These empirical p-values are based on 50 random masks per molecule and 10 molecules per endpoint, so they are coarse diagnostic quantities rather than confirmatory significance tests.

The mean faithfulness gap increased with the masked fraction, from 0.19 logits at 0.05 to 0.71 at 0.30 (Table 7). This pattern is consistent with the model relying more strongly on high-attribution atoms than on random atoms, while additional attributed signal is captured as a larger fraction is masked. Comparisons with endpoint MCC are exploratory because only 11 endpoints and 10 molecules per endpoint were analysed; no correlation or mechanistic localisation claim is inferred from rank ordering alone.

**Table 6: SHAP-Guided Occlusion Faithfulness at Mask Fraction 0.20.** Values are per-task means across 10 evaluated molecules. The SHAP and random columns are mean logit drops after masking high-attribution or size-matched random heavy atoms, respectively. The empirical p-value is a coarse within-molecule randomisation diagnostic, not a confirmatory hypothesis test.

| Endpoint | $\bar{\Delta}_{SHAP}$ | $\bar{\Delta}_{rand}$ | Faithfulness gap | Median $\hat{p}_{emp}$ |
|---|---|---|---|---|
| SR-p53 | 0.97 | 0.09 | 0.88 | 0.06 |
| SR-MMP | 0.87 | 0.06 | 0.81 | 0.04 |
| SR-HSE | 0.65 | 0.03 | 0.61 | 0.04 |
| NR-ER-LBD | 0.65 | 0.09 | 0.56 | 0.04 |
| Ames | 0.63 | 0.09 | 0.53 | 0.10 |
| NR-AhR | 0.62 | 0.11 | 0.51 | 0.02 |
| NR-Aromatase | 0.44 | -0.05 | 0.49 | 0.04 |
| SR-ARE | 0.47 | -0.01 | 0.48 | 0.06 |
| hERG_Karim | 0.36 | 0.01 | 0.34 | 0.16 |
| LD50_Zhu | 0.24 | -0.01 | 0.25 | 0.08 |
| NR-ER | 0.13 | 0.01 | 0.12 | 0.10 |
| **Macro (11)** | **0.55** | **0.04** | **0.51** | **0.06** |

**Table 7: Faithfulness Audit Across Mask Fractions.** Cross-task means are shown for the four evaluated mask fractions. Increasing score-drop differences describe sensitivity to progressively larger attributed regions; they do not establish biological mechanism.

| Mask fraction | Mean $\bar{\Delta}_{SHAP}$ | Mean $\bar{\Delta}_{rand}$ | Mean gain | Median $\hat{p}_{emp}$ across tasks | Tasks with $\hat{p}_{emp} \leq 0.10$ |
|---|---|---|---|---|---|
| 0.05 | 0.20 | 0.01 | 0.19 | 0.10 | 6 / 11 |
| 0.10 | 0.32 | 0.02 | 0.30 | 0.14 | 2 / 11 |
| 0.20 | 0.55 | 0.04 | 0.51 | 0.06 | 10 / 11 |
| 0.30 | 0.66 | -0.05 | 0.71 | 0.04 | 10 / 11 |

## 4.8. Counterfactually Validated Toxicophores

The consensus-motif analysis applied the prespecified seed, expansion, clustering, and counterfactual-occlusion rules described in Section 2.8 to positive test examples. Cluster ranking used the fraction of tested occurrences passing all three occurrence-level criteria, mean probability drop, difference from random controls, and prevalence. Importantly, the label 'counterfactual_validated' in the output table means that a cluster contained at least one passing occurrence; it does not mean that every occurrence or the cluster-average statistics passed each threshold.

The pipeline reported 49 consensus clusters across the 11 endpoints. Forty-four clusters contained at least one occurrence that passed the three counterfactual criteria, whereas five were retained as SHAP-consensus clusters with no passing occurrence. The number of reported clusters varied strongly by endpoint and by the number of labelled positive test molecules. Consequently, cluster counts should not be compared as endpoint performance measures. NR-Aromatase, NR-ER-LBD, and SR-HSE produced only exploratory SHAP-consensus clusters under the configured thresholds. The top-ranked reported motif for each endpoint is summarised in Table 8.

The catalogue contains several chemically recognisable fragments, including phenolic motifs for SR-MMP, basic amine-containing motifs for hERG_Karim, aromatic environments for NR-AhR, aromatic amine and nitroso fragments for Ames, and a phenolic fragment for NR-ER. Some of these are consistent with established structural-alert literature [5, 101, 102], but the fragments are short and the model also uses global molecular features. Their recovery therefore supports chemical plausibility only at the hypothesis-generation level. Representative parent molecules and occurrence-level statistics are required to interpret each fragment in context. Representative motifs and their parent-molecule contexts are shown in Figure 4.

Across tested occurrences, masking SHAP-seeded motifs reduced the predicted positive-class probability by 0.107 on average, compared with 0.015 for size-matched random subgraphs, a mean difference of 0.091. These values support within-model faithfulness of the selected subgraphs relative to the configured random control. They do not establish biological causality, chemical-deletion effects, or experimental toxicophores.

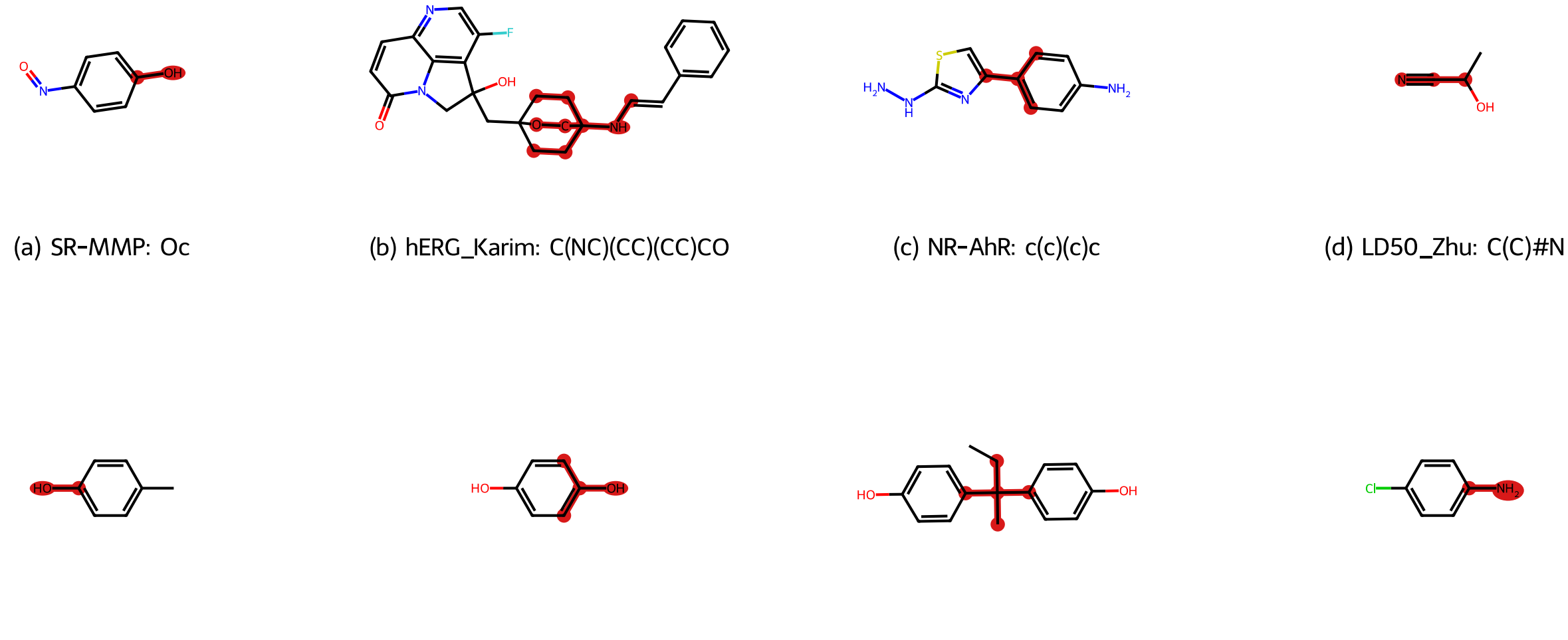

**Figure 4: Counterfactually Validated Consensus Toxicophores per Endpoint.** Each panel renders the top-ranked consensus motif (by validated fraction → mean $\Delta_{true}$ → prevalence) embedded in a representative test-set positive molecule that contains it. Panels: (a) SR-MMP—phenol; (b) hERG_Karim—tertiary β-aminoalcohol; (c) NR-AhR—tetra-substituted aromatic; (d) LD50_Zhu—alkyl nitrile; (e) NR-ER—phenol; (f) SR-ARE—ortho-substituted phenol; (g) SR-p53—geminal diaryl-dimethyl; (h) Ames—aromatic amine. Highlighted atoms (red) are the SHAP-seeded fragment members; un-highlighted atoms are the molecular context.

Several reported fragments are one- or two-atom environments, such as aromatic amine, phenol, carbonyl, and nitroso motifs. This follows from stopping expansion at the smallest radius capturing 35% of positive attribution mass. Short canonical SMILES can be chemically nonspecific; the catalogue must therefore be interpreted together with representative parent molecules, prevalence, validated fraction, and occurrence-level controls. Generic aromatic fragments are retained for transparency when they pass the algorithmic criteria, but they should not be presented as uniquely identifying an endpoint mechanism. The complete 49-cluster catalogue and occurrence-level files are provided as Supplementary Data.

**Table 8: Top-Ranked Reported Consensus Motifs by Endpoint.** A cluster is labelled counterfactually supported when at least one tested occurrence passes all three occurrence-level criteria. Validated fraction is the proportion of tested occurrences that pass. Mean values can therefore lie outside an individual-occurrence threshold. Dagger-marked endpoints produced SHAP-consensus clusters only.

| Endpoint | Top Motif | Validated Fraction | Mean $\Delta_{true}$ | Mean $\hat{p}_{emp}$ | Count | Chemistry |
|---|---|---|---|---|---|---|
| SR-MMP | Phenol (ArOH) | 0.75 | 0.47 | 0.07 | 4 | Mitochondrial uncoupler pharmacophore |

| Endpoint | Top Motif | Validated Fraction | Mean $\Delta_{true}$ | Mean $\hat{p}_{emp}$ | Count | Chemistry |
|---|---|---|---|---|---|---|
| hERG_Karim | *N*-methyl tertiary aminoalcohol | 0.83 | 0.33 | 0.06 | 12 | Tertiary cyclic/open-chain amine; hERG pharmacophore |
| NR-AhR | Trisubstituted aromatic | 0.60 | 0.30 | 0.11 | 5 | Generic aromatic environment; interpret in parent-molecule context |
| LD50_Zhu | Alkyl nitrile (R-C≡N) | 0.67 | 0.22 | 0.05 | 3 | Nitrile-containing motif; mechanism unresolved |
| NR-ER | Phenol (ArOH) | 0.33 | 0.22 | 0.05 | 3 | Oestrogen-receptor phenolic pharmacophore |
| Ames | Aromatic amine ($ArNH_2$) | 0.50 | 0.13 | 0.20 | 4 | Primary aniline-type mutagenicity alert |
| SR-ARE | Ortho-substituted phenol | 0.33 | 0.10 | 0.27 | 3 | Phenolic motif associated with the model output |
| SR-p53 | Geminal diaryl-dimethyl (quaternary C) | 0.33 | 0.06 | 0.38 | 3 | Generic diaryl-carbon motif; mechanism unresolved |
| NR-Aromatase† | — | — | — | — | — | Exploratory only |
| NR-ER-LBD† | — | — | — | — | — | Exploratory only |
| SR-HSE† | — | — | — | — | — | Exploratory only |

## 5. Discussion

ToxLens addresses a practical limitation of molecular-toxicity benchmarking: discrimination alone is insufficient to establish utility for compound triage [27, 33]. Structural separation, applicability-domain behaviour, probability quality, calibrated abstention, and model interpretation provide complementary evidence [28, 31, 41, 44, 56, 79-81]. The contribution of ToxLens is the integration and auditable evaluation of these components in one multi-task workflow.

On the leakage-controlled 11-endpoint panel, the five-seed ensemble achieved macro MCC 0.44, macro AUROC 0.83, and macro AUPRC 0.58. It exceeded four ECFP4-based shallow baselines on all endpoints under the same split and validation-threshold protocol. This demonstrates an advantage over those specific baselines on the chosen partition, not superiority over all published toxicity models. The fixed UMAP-HDBSCAN split was designed to reduce close-analogue leakage, but conclusions remain conditional on one partition and one endpoint panel.

The ablation study showed that removing the global pathway produced the weakest tested configuration, whereas the tested GCMI and FiLM mechanisms did not improve on late concatenation. The defensible conclusion is empirical: simple late concatenation was the strongest tested design for this panel. The proposed explanation - that mid-trunk modulation may suppress complementary graph information - remains a hypothesis because the ablation did not directly measure information loss or modality collapse.

The conformal-style layer changed how the predictions are reported by allowing a two-label set when both classes satisfy the calibrated threshold. Efficiency varied substantially by endpoint, from 20.40% singleton predictions for LD50_Zhu to 100% for NR-ER-LBD. This abstention mechanism can support triage, but reuse of the validation fold for checkpoint selection, temperature scaling, and set calibration precludes an exact split-conformal finite-sample guarantee. The nominal 95% level is therefore a target operating level, not a guarantee that every singleton prediction is correct; prospective calibration drift was not evaluated.

Applicability-domain stratification showed better aggregate discrimination and calibration among molecules more similar to the training set. The trend was not monotonic for every endpoint, particularly where quartile-specific positive counts were small. Similarity should therefore be used as a cohort-level audit and supplementary molecule-level warning, not as a deterministic acceptance rule.

External retraining experiments showed competitive performance on the canonical Tox21 Challenge and selected TDC ADMET folds, but they did not establish universal state-of-the-art performance. These experiments assess whether the architecture and featurisation can be retrained under externally specified protocols; they are not zero-shot tests of the fitted 11-endpoint model. This distinction is essential because changing the endpoint changes the fitted task heads and optimisation problem even when the architectural template is fixed.

The interpretability workflow linked atom-level GradientSHAP maps [15, 88] to graph-feature occlusion controls and recurring motif mining. Larger score drops under SHAP-guided than random masking support within-model faithfulness, while chemically familiar fragments provide plausible hypotheses. Neither result demonstrates biological mechanism. The consensus motifs are computational structural alerts that require independent chemical review and experimental validation before use as toxicophores.

Taken together, the results support ToxLens as a reproducible decision-support framework that combines predictive modelling with leakage auditing, uncertainty reporting, applicability-domain analysis, and model-level interpretation. The evidence is strongest for comparison with the reported shallow baselines on the fixed in-house split and more limited for claims of external superiority or mechanism.

# 6. Limitations

Four limitations constrain the conclusions. First, the in-house comparison uses one UMAP-HDBSCAN partition; the five ensemble seeds quantify optimisation variability but not split variability. Second, external benchmarks retrain the architectural template on different task panels and therefore test protocol-level transfer rather than external validity of the fitted 11-endpoint checkpoint. ToxLens is competitive on these folds but does not uniformly exceed the reported reference values. Third, the prediction-set procedure reuses the validation fold for checkpoint selection, temperature scaling, and calibration, so the exact finite-sample split-conformal guarantee does not apply; efficiency and empirical coverage may also change under prospective distribution shift. Fourth, SHAP, feature occlusion, and consensus clustering are model-level analyses. The motif catalogue includes short, sometimes generic fragments, and no prospective experimental validation was performed. In addition, the 200-dimensional PubChem block was zero in the reported checkpoint because the cache was absent, so no conclusion can be drawn about PubChem bioactivity features.

# 7. Future Directions

Future work should evaluate repeated chemistry-aware splits and prospective temporal or external cohorts, recalibrate conformal prediction under measured deployment shift, and test model-derived motifs experimentally. A multi-seed Tox21 Challenge evaluation would reduce the present ensemble-depth asymmetry with published DeepTox results. Assay-condition metadata, including cell line, exposure duration, and readout type, may also support larger multi-assay models, but this extension should be evaluated for negative transfer rather than assumed beneficial.

# 8. Conclusion

ToxLens combines an 11-task graph-plus-global classifier with leakage-aware partitioning, validation-based threshold selection, conformal-style prediction sets, applicability-domain analysis, fixed-split external retraining, and counterfactually controlled model interpretation. On the in-house held-out split, the five-seed ensemble achieved macro MCC 0.44, macro AUROC 0.83, and macro AUPRC 0.58, exceeding the reported ECFP4 shallow baselines on all endpoints. External results were competitive but not uniformly state of the art. The study therefore supports a broader evaluation framework for computational toxicity triage while identifying clear limits on split generality, prediction-set calibration, and mechanistic interpretation.

## Supplementary Information

The following supplementary materials accompany this manuscript.

- Supp. Table 1 (per-task class-balance counts, thresholds, and conformal quantiles)
- Supp. Table 2 (per-task shallow-baseline vs. ensemble MCC)
- Supp. Table 3 (data-partition strategy comparison)
- Supp. Table 4 (applicability-domain stratified performance)
- Supp. Table 5 (extended Tox21 Challenge per-task metrics with EF@1% and BEDROC)
- Supp. Table 6 (per-task ranking-quality metrics on the in-house test fold).
- Supp. Table 7 (TDC ADMET retraining: per-seed AUROC, development/test overlap audit, and published-comparator values)
- External benchmark provenance: PROVENANCE.json (exact fold SHA-256 hashes), all_benchmark_summary.csv, all_published_method_comparisons.csv

## Abbreviations

AUPRC, area under the precision-recall curve; AUROC, area under the receiver operating characteristic curve; BCE, binary cross-entropy; CI, confidence interval; DBSCAN, density-based spatial clustering of applications with noise; ECFP4, extended-connectivity fingerprint with radius 2; FiLM, feature-wise linear modulation; GCMI, gated cross-modality interaction; GINE, graph isomorphism network with edge features; HDBSCAN, hierarchical density-based spatial clustering of applications with noise; MCC, Matthews correlation coefficient; MC, Monte Carlo; MMFF, Merck molecular force field; NR, nuclear receptor; OOD, out-of-distribution; PAINS, pan-assay interference compounds; QSAR, quantitative structure-activity relationship; SHAP, Shapley additive explanations; SMARTS, SMILES arbitrary target specification; SR, stress response; TDC, Therapeutics Data Commons; UMAP, uniform manifold approximation and projection.

## Declarations

## Availability of data and materials

The endpoint data were obtained from the Therapeutics Data Commons toxicity tasks (Ames, LD50_Zhu, and Tox21 labels) and the published hERG_Karim dataset [7]. The external Tox21 Challenge benchmark used the canonical DeepTox archive [11]. Source code, fixed split assignments, selected checkpoints, the software environment, benchmark-provenance records, and machine-readable results are provided at https://github.com/Magnushst/toxlens.

## Authors' contributions

M.H.S. curated the data, designed and implemented the framework, conducted the experiments, analysed the results, and drafted the manuscript. A.G.C.d.S. and D.B.A. supervised the work and revised the manuscript. All authors read and approved the final manuscript.

## Funding

D.B.A. is supported by an NHMRC Investigator (GRNT2041888). Research supported by the NVIDIA Academic Grant Program.

## Competing interests

The authors declare no competing interests.

## Ethics approval and consent to participate

Not applicable.

## Consent for publication

Not applicable.

## Acknowledgements

Not applicable.

# Supplementary materials

## 1. Supplementary Data

### 1.1. Data Partitioning Comparison

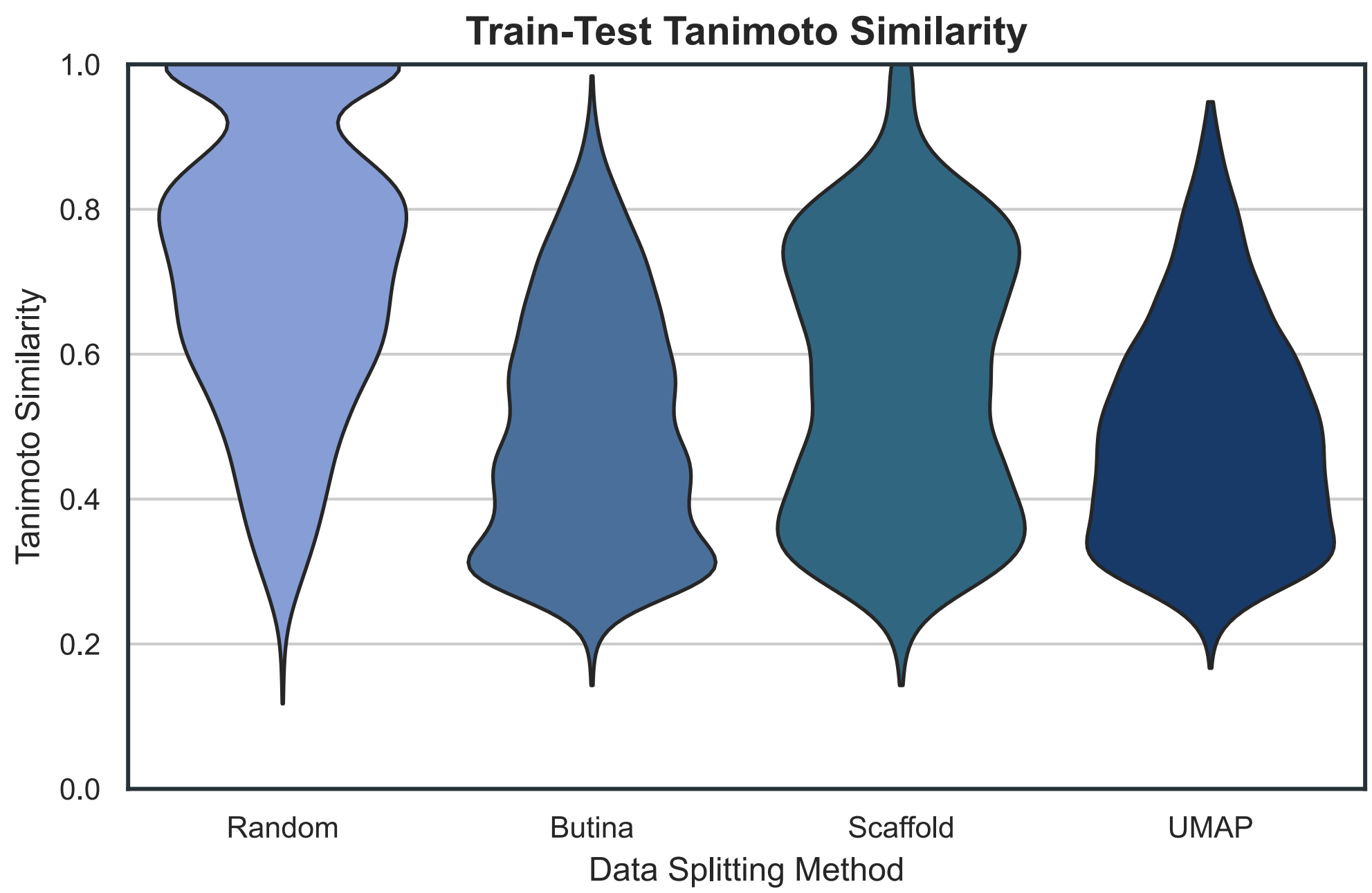

**Supplementary Figure 1: Train-test Tanimoto-similarity distributions for Random, Butina, Bemis-Murcko scaffold, and UMAP-HDBSCAN partitions.** For each test molecule, similarity is measured against its nearest training-set neighbour; lower distributions indicate stronger structural separation. Complementary scaffold-novelty, property-balance, and aggregate out-of-distribution statistics are reported in Supplementary Table 3.

### 1.2. Class Balance and Loss Weighting

The 11 endpoints differed substantially in training-fold class balance. The most imbalanced endpoint was NR-Aromatase (194 positives and 3,969 negatives), followed by NR-ER-LBD (246 and 4,716) and SR-HSE (271 and 4,388). Ames (2,508 and 2,319) and hERG_Karim (4,377 and 4,541) were approximately balanced. Although the implementation can compute inverse-frequency class weights, the retained checkpoint used class-weight power zero and therefore unweighted binary cross-entropy. Fixed endpoint-emphasis weights and validation-selected decision thresholds remained active. Counts, thresholds, and conformal quantiles are reported in Supplementary Table 1.

**Supplementary Table 1: Per-Task Training and Validation Counts, Decision Thresholds, and Conformal Quantiles.** Counts follow duplicate resolution and the final UMAP-HDBSCAN split.

| Endpoint | Train pos / neg | Validation pos / neg | Class ratio (train) | Validation threshold | Conformal $\hat{q}$ |
|---|---|---|---|---|---|
| Ames | 2 508 / 2 319 | 437 / 347 | 0.9 : 1 | 0.77 | 0.81 |
| LD50_Zhu | 2 518 / 3 152 | 300 / 394 | 1.3 : 1 | 0.37 | 0.80 |
| hERG_Karim | 4 377 / 4 541 | 549 / 511 | 1.0 : 1 | 0.65 | 0.78 |
| NR-AhR | 564 / 4 100 | 94 / 498 | 7.3 : 1 | 0.51 | 0.62 |
| NR-Aromatase | 194 / 3 969 | 25 / 491 | 20.4 : 1 | 0.77 | 0.55 |
| NR-ER | 561 / 3 866 | 79 / 485 | 6.9 : 1 | 0.30 | 0.88 |
| NR-ER-LBD | 246 / 4 716 | 22 / 601 | 19.2 : 1 | 0.29 | 0.29 |
| SR-ARE | 655 / 3 546 | 92 / 439 | 5.4 : 1 | 0.16 | 0.91 |
| SR-HSE | 271 / 4 388 | 45 / 553 | 16.3 : 1 | 0.10 | 0.82 |
| SR-MMP | 609 / 3 569 | 72 / 420 | 5.8 : 1 | 0.06 | 0.84 |
| SR-p53 | 281 / 4 565 | 51 / 563 | 16.3 : 1 | 0.16 | 0.75 |

Inverse-frequency class weights were computed from the training-fold positive and negative counts for diagnostic and implementation purposes, with the positive-to-negative ratio capped at 12.0. In the retained checkpoint, however, the class-weight exponent was zero; both class weights therefore reduced to one, and the fitted objective was unweighted binary cross-entropy with fixed endpoint-emphasis weights. Supplementary Table 1 reports the counts used in this audit.

# 2. Supplementary Methods

## 2.1. Hyperparameter Optimisation

Model selection was restricted to the training and validation folds. The retained checkpoint records the following active configuration: five GINE layers; hidden width 256; learning rate $3 \times 10^{-4}$; weight decay 0.01; batch size 128; hidden dropout 0.20; global-input dropout 0.25; edge-drop probability 0.06; maximum stochastic-depth probability 0.20; final-representation dropout 0.10; cosine-restart period 45; and label-smoothing coefficient 0.0007. It used unweighted binary cross-entropy with fixed endpoint-emphasis weights, no auxiliary-task loss, no graph-only auxiliary loss, no GPS attention, and no mid-trunk fusion. The final checkpoint was selected by validation MCC. Test data were not used for hyperparameter or checkpoint selection.

## 2.2. Checkpoint Selection and Probability Calibration

For the retained single model, logits were converted to probabilities with the sigmoid function. A task-specific threshold was selected on the validation fold by evaluating probabilities on a 0.05-0.95 grid in increments of 0.01 and retaining the threshold with the highest validation MCC. These thresholds were stored and applied unchanged to the test fold.

Temperature scaling was fitted separately for the post-hoc conformal analysis and did not alter AUROC or the headline uncalibrated point-prediction probabilities.

## 2.3. Gated Cross-Modality (ablated comparator)

A gated cross-modal fusion module, termed Gated Cross-Modality Interaction (GCMI), was evaluated as an ablated mechanism for integrating node-level graph representations with the global vector. The final model does not use GCMI because late concatenation achieved higher test MCC in the controlled ablation (Section 4.3). The equations below document only the ablated comparator.

Let $X \in \mathbb{R}^{N \times d_n}$ be the node-feature matrix, where $N$ is the total number of atoms across all graphs in the batch and rows are mapped to their parent graph through the PyTorch Geometric batch index $b: 1, \dots, N \to 1, \dots, B$, and let $G \in \mathbb{R}^{B \times d_g}$ be the per-graph global-descriptor matrix. GCMI broadcasts the global vector to the node level, $\tilde{G}_i = G_{b(i)}$, and computes:

$$g_i = \sigma\left(W_g \tilde{G}_i + b_g\right) \in (0,1)^{d_n} \quad (S2.1)$$

$$s_i = \tanh\left(W_s\left[(g_i \odot x_i) \parallel \tilde{G}_i\right] + b_s\right) \in \mathbb{R}^{d_o} \quad (S2.2)$$

$$r_i = W_r x_i + b_r \in \mathbb{R}^{d_o} \quad (S2.3)$$

$$y_i = \mathrm{LayerNorm}(r_i + s_i) \quad (S2.4)$$

The contextual gate reweights node channels as a function of the global representation, and the synergy term models nonlinear interactions between gated node features and global context. An ungated residual connection preserves a direct structural path. Gate statistics were recorded only for the GCMI comparator; they are not produced by the retained no-fusion model.

## 2.4. Classification Loss: Masked Binary Cross-Entropy with Fixed Task Emphasis

Multi-task binary classification is trained under a masked, macro-averaged binary cross-entropy (BCE) objective with label smoothing and fixed per-task emphasis weights. The loss exposes no learnable per-task scalars; task balancing is governed entirely by fixed weights set prior to training, and class imbalance is addressed primarily through validation-based per-task threshold calibration (Supplementary §2.2) rather than through the loss.

Let $z_{i,t}$ denote the logit predicted for sample $i$ on task $t$, with label $y_{i,t} \in \{0, 1, \mathrm{NaN}, -1\}$ (NaN and $-1$ encode missing entries under the masked multi-task setup), $p_{i,t} = \sigma(z_{i,t})$, and $V_t = \left\{i: y_{i,t} \in \{0,1\}\right\}$ the set of valid labels for task $t$ in the current batch. Each valid label is smoothed as $\tilde{y}_{i,t} = y_{i,t}(1-\epsilon) + \frac{1}{2}\epsilon$, with $\epsilon = 7 \times 10^{-4}$, which mildly attenuates overconfident logits. The per-task loss is the class-weighted mean BCE over valid labels,

$$L_t = \frac{1}{|V_t|} \sum_{i \in V_t} \alpha_{i,t} \ \mathrm{BCE}\big(z_{i,t}, \ \tilde{y}_{i,t}\big) \tag{S2.5}$$

$$\alpha_{i,t} = \begin{cases} w_t^+ & y_{i,t} = 1 \\ w_t^- & y_{i,t} = 0 \end{cases}$$

where $\{w_t^+, w_t^-\}$ are fixed (non-learned) class-balance buffers. These buffers are computed from training-fold counts as $w_t^c = N_t/(2\, n_t^c)$ with the positive-to-negative weight ratio capped at 12.0; however, in the active configuration the class-weight exponent is set to zero, which reduces both buffers to unity ($w_t^+ = w_t^- = 1$) and recovers unweighted BCE. The total objective is the task-emphasis-weighted macro average over the tasks that carry at least one valid label in the batch,

$$L = \frac{\sum_{t=1}^{T} \mathbb{1}\,[\,|V_t| > 0\,]\ \beta_t\, L_t}{\sum_{t=1}^{T} \mathbb{1}\,[\,|V_t| > 0\,]\ \beta_t + \varepsilon} \tag{S2.6}$$

Fixed endpoint-emphasis weights were set above unity for selected primary tasks: 1.35 for NR-ER and NR-ER-LBD; 1.30 for SR-p53; 1.25 for hERG_Karim and NR-Aromatase; 1.20 for NR-AhR and SR-MMP; and 1.10 for Ames and LD50_Zhu. All other tasks used weight 1.0. Graph-only auxiliary supervision and auxiliary-transfer losses were implemented but had weights of zero in the retained checkpoint, so only the primary masked objective contributed to the reported model.

This formulation differs from learnable-task-weighting schemes in that it exposes no trainable scalars per task: imbalance and task prioritisation are handled through the fixed class buffers (inactive in the reported configuration), the fixed task-emphasis weights, label smoothing, and validation-based per-task thresholding.

## 2.5. MC Dropout and Conformal-Style Prediction Sets for Binary Classification

A marginal error rate alpha = 0.05 was targeted using the held-out validation fold. Thirty stochastic forward passes were run with dropout modules active and normalisation layers in evaluation mode. Temperature-scaled per-task predictive means were used for conformal calibration, and the standard deviations across passes were retained as separate epistemic-variability diagnostics.

For endpoint t, the nonconformity score was 1 - p for a positive validation label and p for a negative label, where p is the temperature-scaled MC predictive mean. With n valid calibration labels, the higher empirical quantile was evaluated at min(1, ceil((n + 1)(1 - alpha))/n). Missing labels were excluded independently for each endpoint.

At inference, class 0 was included when p was no greater than the calibrated quantile and class 1 when 1 - p was no greater than the quantile. Two-label sets were flagged as uncertain. Empty sets were collapsed to the probability argmax by the implementation. Because the same validation fold informed checkpoint selection, temperature scaling, and calibration, these are conformal-style sets with $\alpha = 0.05$ as a target operating level; an exact split-conformal coverage guarantee is not claimed. MC-dropout standard deviation is reported alongside the set but does not enter its construction.

## 3. Supplementary Results

**Supplementary Table 2: Test-Set MCC of Shallow Baselines (Random Forest, XGBoost, MLP, SVM) Against the Five-Seed ToxLens Ensemble Across Eleven Endpoints.** All baselines use 1024-bit ECFP4 and share the UMAP-HDBSCAN split and per-task validation-MCC threshold calibration with the ToxLens model. Values are computed at each model's validation-optimal threshold.

| Model | Endpoints | | | | | | | | | | |
|---|---|---|---|---|---|---|---|---|---|---|---|
| | **Ames** | **LD50_Zhu** | **hERG_Karim** | **NR-AhR** | **NR-Aromatase** | **NR-ER** | **NR-ER-LBD** | **SR-ARE** | **SR-HSE** | **SR-MMP** | **SR-p53** |
| RF | 0.55 | 0.39 | 0.25 | 0.41 | 0.08 | 0.29 | 0.47 | 0.29 | 0.14 | 0.42 | 0.20 |
| XGBoost | 0.50 | 0.32 | 0.29 | 0.36 | 0.10 | 0.32 | 0.45 | 0.21 | −0.01 | 0.40 | 0.23 |
| MLP | 0.51 | 0.32 | 0.24 | 0.42 | 0.10 | 0.15 | 0.24 | 0.19 | 0.03 | 0.32 | 0.21 |
| SVM | 0.36 | 0.28 | 0.22 | 0.13 | 0.16 | 0.07 | 0.17 | 0.11 | 0.13 | 0.31 | 0.21 |
| **ToxLens Ensemble** | **0.59** | **0.47** | **0.36** | **0.45** | **0.23** | **0.47** | **0.59** | **0.39** | **0.29** | **0.61** | **0.36** |

**Supplementary Table 3: Comparison of Data-Partition Strategies.** Random and Butina baselines are reported alongside Bemis–Murcko scaffold and the chosen UMAP-HDBSCAN spatial split. Median Tc is the median Tanimoto similarity between each test molecule and its nearest training neighbour (lower = stronger separation). Novel scaffolds % is the fraction of test-set Bemis–Murcko scaffolds unseen during training. Structural novelty, property balance, and OOD utility are composite scores in [0, 1]; the OOD utility score combines the first two with a label-distribution distance penalty. The UMAP-HDBSCAN split attains the highest OOD utility while preserving non-degenerate property balance, which Scaffold splitting sacrifices.

| Split | Median Tc | Novel Scaffolds (%) | Structural Novelty | Property Balance | OOD utility |
|---|---|---|---|---|---|

| Random | 0.73 | 37.9 | 0.27 | 0.35 | 0.33 |
|---|---|---|---|---|---|
| Butina | 0.47 | 75.1 | 0.53 | 0.13 | 0.37 |
| Scaffold | 0.56 | 100 | 0.44 | 0.05 | 0.29 |
| UMAP | 0.47 | 68.7 | 0.53 | 0.15 | 0.38 |

**Supplementary Table 4: Applicability-Domain Stratified Held-Out Test Fold Performance.** Test molecules are binned into quartiles by their maximum Tanimoto similarity to the training set (Q1 = most distant, Q4 = most similar; n ≈ 742-748 per quartile per task on the 2,970-molecule held-out test split). The expected monotonic improvement from Q1 to Q4 is observed on the macro mean (mean AUROC rises 0.77 → 0.82 → 0.82 → 0.86 and mean MCC rises 0.28 → 0.35 → 0.41 → 0.44 across the quartiles shown; the corresponding mean expected calibration error, computed separately, falls monotonically from 0.13 (Q1) to 0.07 (Q4)), confirming that model confidence tracks structural proximity to the training distribution. Per-task behaviour is heterogeneous—NR-ER-LBD shows the cleanest monotonic improvement (AUROC 0.71 → 0.81 → 0.95 → 0.96; MCC 0.00 → 0.46 → 0.74 → 0.78), Ames is essentially flat across quartiles (it is the largest and most chemically diverse endpoint, so all quartiles are already within-distribution), and a handful of endpoints (SR-HSE, NR-AhR) reverse on individual quartiles where positive-class counts are small enough to inject sampling noise into the bootstrap CI. The conformal layer (Table 3) converts the cohort-level confidence-vs-novelty relationship into an explicit per-task abstention rate, surfacing structural novelty to downstream users at inference rather than at audit time.

| Endpoint | Q1 AUROC | Q1 MCC | Q2 AUROC | Q2 MCC | Q3 AUROC | Q3 MCC | Q4 AUROC | Q4 MCC |
|---|---|---|---|---|---|---|---|---|
| Ames | 0.84 | 0.51 | 0.87 | 0.49 | 0.84 | 0.50 | 0.87 | 0.56 |
| LD50_Zhu | 0.75 | 0.41 | 0.74 | 0.36 | 0.73 | 0.37 | 0.89 | 0.57 |
| hERG_Karim | 0.65 | 0.24 | 0.78 | 0.32 | 0.76 | 0.35 | 0.87 | 0.49 |
| NR-AhR | 0.94 | 0.64 | 0.82 | 0.28 | 0.86 | 0.64 | 0.88 | 0.47 |
| NR-Aromatase | 0.82 | 0.37 | 0.77 | 0.43 | 0.77 | −0.02 | 0.89 | 0.29 |
| NR-ER | 0.52 | −0.05 | 0.84 | 0.29 | 0.80 | 0.39 | 0.78 | 0.45 |
| NR-ER-LBD | 0.71 | 0.00 | 0.81 | 0.46 | 0.95 | 0.74 | 0.96 | 0.78 |
| SR-ARE | 0.69 | 0.04 | 0.78 | 0.26 | 0.79 | 0.31 | 0.78 | 0.38 |
| SR-HSE | 0.78 | 0.31 | 0.89 | 0.25 | 0.79 | 0.13 | 0.75 | −0.01 |
| SR-MMP | 0.96 | 0.40 | 0.89 | 0.52 | 0.89 | 0.58 | 0.91 | 0.50 |
| SR-p53 | 0.77 | 0.24 | 0.81 | 0.19 | 0.88 | 0.49 | 0.87 | 0.31 |
| **Macro (11)** | **0.77** | **0.28** | **0.82** | **0.35** | **0.82** | **0.41** | **0.86** | **0.44** |

**Supplementary Table 5: Extended Tox21 Challenge final-test per-task metrics including EF@1% and BEDROC.** Single-seed ToxLens evaluation on the canonical DeepTox tox21_labels_test.csv.gz partition, at the per-task validation-optimal probability threshold.

CIs are 1,000-iteration percentile bootstrap. Rows are sorted by descending AUROC within the reportable-primary set, with excluded endpoints appended.

| Endpoint | AUROC | AUPRC | MCC | EF@1% | BEDROC (α=20) |
|---|---|---|---|---|---|
| SR-MMP | 0.95 [0.91, 0.97] | 0.63 [0.53, 0.77] | 0.55 [0.43, 0.67] | 6.22 | 0.70 |
| NR-AhR | 0.91 [0.87, 0.94] | 0.63 [0.53, 0.74] | 0.56 [0.47, 0.65] | 7.34 | 0.72 |
| SR-p53 | 0.84 [0.78, 0.89] | 0.31 [0.21, 0.44] | 0.20 [0.06, 0.34] | 6.58 | 0.40 |
| SR-HSE | 0.81 [0.70, 0.91] | 0.31 [0.17, 0.55] | 0.38 [0.16, 0.59] | 20.68 | 0.47 |
| NR-Aromatase | 0.79 [0.71, 0.86] | 0.24 [0.17, 0.37] | 0.30 [0.18, 0.42] | 4.51 | 0.33 |
| NR-ER | 0.79 [0.72, 0.86] | 0.49 [0.37, 0.62] | 0.45 [0.31, 0.58] | 10.30 | 0.61 |
| SR-ARE | 0.79 [0.74, 0.84] | 0.44 [0.37, 0.55] | 0.37 [0.29, 0.46] | 3.01 | 0.57 |
| SR-ATAD5 | 0.78 [0.70, 0.85] | 0.33 [0.20, 0.48] | 0.37 [0.18, 0.53] | 14.36 | 0.43 |
| NR-ER-LBD | 0.77 [0.66, 0.86] | 0.23 [0.08, 0.41] | 0.38 [0.00, 0.54] | 14.98 | 0.29 |
| NR-PPAR-gamma | 0.75 [0.65, 0.84] | 0.17 [0.11, 0.30] | 0.19 [0.05, 0.33] | 5.56 | 0.30 |
| **10-task macro** | **0.82** | **0.38** | **0.37** | **9.35** | **0.48** |
| NR-AR (excluded) | 0.85 [0.71, 0.97] | 0.37 [0.17, 0.64] | 0.21 [-0.01, 0.46] | 16.28 | 0.56 |
| NR-AR-LBD (excluded) | 0.79 [0.63, 0.91] | 0.08 [0.03, 0.33] | 0.08 [-0.02, 0.27] | 12.10 | 0.22 |
| **12-task macro (all)** | **0.82 [0.73, 0.89]** | **0.35 [0.24, 0.52]** | **0.34 [0.17, 0.49]** | **10.16** | **0.47** |

**Supplementary Table 6: Per-Task Ranking-Quality Metrics on the UMAP-HDBSCAN Hold-Out Test Fold.** BEDROC at α = 20; EF@k% is the enrichment factor at the top k % of ranked predictions.

| Endpoint | BEDROC (α=20) | EF@1% | EF@5% | EF@10% | Cohen's κ |
|---|---|---|---|---|---|
| Ames | 0.94 | 1.36 | 1.45 | 1.45 | 0.54 |
| LD50_Zhu | 0.78 | 2.11 | 1.88 | 1.84 | 0.42 |
| hERG_Karim | 0.84 | 1.68 | 1.66 | 1.68 | 0.39 |
| NR-AhR | 0.74 | 7.58 | 6.32 | 4.76 | 0.45 |

| Endpoint | BEDROC (α=20) | EF@1% | EF@5% | EF@10% | Cohen's κ |
|---|---|---|---|---|---|
| NR-Aromatase | 0.37 | 10.65 | 5.68 | 3.19 | 0.25 |
| NR-ER | 0.59 | 9.66 | 5.52 | 4.04 | 0.35 |
| NR-ER-LBD | 0.68 | 21.83 | 10.23 | 6.48 | 0.54 |
| SR-ARE | 0.64 | 3.95 | 3.34 | 2.63 | 0.29 |
| SR-HSE | 0.36 | 7.73 | 5.60 | 4.80 | 0.21 |
| SR-MMP | 0.89 | 3.43 | 3.16 | 3.03 | 0.62 |
| SR-p53 | 0.45 | 4.41 | 4.41 | 4.19 | 0.33 |
| **Macro (11)** | **0.66** | **6.76** | **4.48** | **3.46** | **0.40** |

**Supplementary Table 7: TDC ADMET Retraining Comparison on Fixed Benchmark Folds.** ToxLens AUROC is the mean ± SD across five 90:10 development re-splits, each retrained from scratch with checkpoint selection on validation MCC and evaluated once on the untouched benchmark test fold. n(dev)/n(test) are the benchmark fold sizes; dev/test overlap is the count of canonical molecules shared between the development and test folds (an inherent property of the benchmark). clean_claim is False where inherent dev/test overlap exists. Comparator AUROCs are TDC leaderboard point estimates associated with the named published architectures, not reruns by this study.

| Benchmark | ToxLens AUROC | MCC | n(dev)/n(test) | Dev/Test overlap | clean_claim | TDC reference comparator |
|---|---|---|---|---|---|---|
| AMES | 0.83 ± 0.01 | 0.52 | 5821/1457 | 0 | True | Chemprop-RDKit 0.850 |
| DILI | 0.89 ± 0.02 | 0.62 | 379/96 | 0 | True | Chemprop 0.899 |
| hERG† | 0.80 ± 0.05 | 0.41 | 523/132 | 6 | False | DeepPurpose 0.841 |

Per-comparator point-estimate deltas (ToxLens − TDC reference): AMES vs Chemprop-RDKit −0.019, CMPNN −0.012, AttentiveFP +0.017; DILI vs Chemprop −0.012, Chemprop-RDKit −0.000, AttentiveFP +0.001; hERG vs DeepPurpose −0.036, Chemprop-RDKit −0.035, AttentiveFP −0.020.